\documentclass{article}
\usepackage{ijcai18}

\usepackage{times}
\usepackage{xcolor}
\usepackage{color}
\usepackage{soul}
\usepackage[utf8]{inputenc}
\usepackage[small]{caption}
\usepackage[normalem]{ulem}

\usepackage[caption = false]{subfig}

\usepackage{amsmath}
\usepackage{graphicx}
\usepackage[colorinlistoftodos]{todonotes}
\usepackage[autostyle, english = american]{csquotes}

\title{Morphology dictates a robot’s ability to ground crowd-proposed language}

\author{
Zahra Mahoor, 
Jack Felag,
Josh Bongard
\\ 
University of Vermont, Computer Science Department\\
zmahoor@uvm.edu,
jfelag@uvm.edu,
jbongard@uvm.edu
}

\begin{document}

\maketitle

\begin{abstract}
{\color{black}As more robots act in physical proximity to people,} it is essential to ensure they make decisions and execute actions that align with human values. 
{\color{black} To do so, robots need to understand the true intentions behind human-issued commands
}. 
In this paper, we define a safe robot as one that 
{\color{black}receives a natural-language command from humans,
considers an action in response to that command,
and accurately predicts how humans will judge that action if executed in reality.
} 
Our contribution is two-fold: 
First, we introduce a web platform for human users to propose commands to simulated robots. 
The robots receive commands and act based on those proposed commands, and then the users provide positive and/or negative reinforcement. 
Next, we train a critic for each robot to predict the crowd's responses to one of the crowd-proposed commands. 
Second, we show that the morphology of a robot plays a role in the way it grounds language:
The critics show that two of the robots used in the experiment achieve a lower prediction error than the others. 
Thus, those two robots are safer, according to our definition, since they ground the proposed command more accurately\footnote{All source code and data are available at https://github.com/zmahoor/TPR-1.0.git}$^{,}$\footnote{See https://youtu.be/KLH1Vff9GpI for a video of the experiment in operation.}.

\end{abstract}

\section{Introduction}
Consider a home assistant robot asked to fetch a book from a room, but the door is jammed. 
Since the robot cannot open the door, it breaks through the door to retrieve the book. 
Because of this undesired and unexpected result, the robot is reprogrammed with a new utility function that penalizes actions that cause the robot to break the door. 
After reprogramming, assume that the robot is then asked to fetch a life-saving medicine from a room but again faces a jammed door. 
This time the robot returns empty-handed instead of breaking through the door and retrieving the medicine. 
The outcomes of both scenarios are unsatisfactory because 
{\color{black}of the failure
}
to communicate our true desires to an AI {\color{black} using just an objective function}. 
These scenarios introduce an important challenge in AI safety called ``perverse instantiation" \cite{bostrom} or ``value misalignment", in which an AI's values do not align with human values (\cite{DBLP:journals/corr/Hadfield-Menell16}).

\cite{yudkowsky} claims that programming our wishes or desires into an AI is not adequate to address this challenge. 
Instead, we should develop methods to enable an AI to learn our intentions and act based on those intentions. 
In other words, we require an AI to infer the intent behind our commands instead of following them verbatim: ``Do as I mean, not as I say.''
One approach to align an AI's values introduced by \cite{DBLP:journals/corr/Hadfield-Menell16} is cooperative inverse reinforcement learning (CIRL), in which a human and a robot play a game so that the robot learns the human’s reward function (i.e. the human's values). 
In this game it is important that the robot is initially unaware and uncertain of the reward function. 
\cite{Christiano} reports a specific instantiation of CIRL: a reinforcement learning agent learns reward functions by receiving feedback from a human in Atari games or locomotion tasks.  However, in CIRL, it is not clear how the robot can align its values with those of a group of people, some of whom may have differing values.

In this work, we define an AI to be safe if it can predict how people will react to a possible action the AI is considering after receiving a command from the same (or different) people. 
For example, a safe robot, after receiving the command ``get the medicine", should predict that breaking down the door to retrieve the medicine would be met by positive reinforcement from any observers, and not breaking it down will be met with negative reinforcement. 
Although this prediction ability for a safe AI is necessary, it is not sufficient. 
For example, if a robot predicts a negative reaction from breaking through a door, it could still choose to 
{\color{black} do so
}.

Following our definition of safety, we here propose a game where a group of people issues a command to a robot. 
The robot acts on the issued command and receives positive and/or negative feedback from the human observers. 
The robot initially might not be able to predict the observers' reaction to its action, but eventually, after many trials, it may learn to predict accurately. 
To be safe in this manner {\color{black}
}, a robot must thus find relationships between the language symbols comprising the issued command, the sensorimotor data it generates as a result of responding to those symbols, and the crowd's responses to its actions.

\cite{harnad} states that a symbol must be coupled with the sensorimotor capacities of an agent in order to be grounded.
This means that the agent should be able to recognize what the symbol refers to, and the agent's sensorimotor interactions with the world as influenced by its body should match the representation of the symbol. 
However, this definition of symbol grounding does not specify a metric determining whether, or how well, a particular grounding has been achieved from a human observer's point of view. 
For instance the robot in the example above has grounded the command ``fetch" in its sensorimotor data but not according to human values. 
Since human values can not be measured precisely, we here define a proxy for them: crowd-based reinforcement of an agent's action in response to a crowd-issued command. 
In our proposed game, an agent must ground the symbols comprising the command in its own sensorimotor data {\color{black}
} and this human feedback. 
Recently \cite{DBLP:journals/corr/ChaplotSPRS17} and \cite{DBLP:journals/corr/HermannHGWFSSCJ17} introduced a reinforcement learning agent that can relate language (symbols) to the world and it own actions.
There, neural networks were trained to receive images of an environment and natural language instruction as input and execute the instruction in a 3D virtual environment.
The two works differ in the way the images of the environment and the given instruction combine, but are the same in that the agents do not have bodies (except for the cameras in the environment) with which to 
{\color{black}influence the kind of sensorimotor data they create
}.
Also, the agents ground symbols without receiving any feedback from humans.  
In contrast, agents in our proposed game have {\color{black} different} bodies and receive feedback from human observers and can thus ground crowd-proposed symbols in their sensorimotor experiences according to human values.

We implement our proposed game in a web platform where a crowd of people can help robots ground symbols in this way. 
In this platform, a robot acts after hearing a crowd-proposed command, and receives back positive and/or negative reinforcement from them. 
We use evolutionary algorithms to evolve robots to obtain increasingly more positive reinforcement from the crowd, and a separate learning algorithm to ground symbols in actions and the crowd feedback. 
We allow the crowd to issue any commands they like to the robots, instead of predefining action words for the robots.

In recent years, crowdsourcing has been employed in robotics for action planning and reasoning, object recognition, and robot design. 
For example, \cite{Breazeal:2013:CHI:3109699.3109704} created a two-player game in which a crowd helps build a set of action plans and reasoning strategies for robots. 
\cite{kent} exploited crowdsourcing to create 3D models of objects that are to be grasped by a robot's hand. 
In \cite{wagy}, the crowd designed the bodies for robots while a search method generated successful gaits for those bodies. 
In our work, we use crowdsourcing to help robots assign meaning to symbols while ensuring those meanings align with human values. 
\cite{anetsberger} also used crowdsourcing to enable robots to ground crowd-proposed commands in their sensorimotor experience and the social response to those actions.
In addition to grounding symbols as in \cite{anetsberger}, we show here that a robot's safety---the amount of value alignment that can be achieved from a given amount of crowd effort---depends on aspects of the robot's morphology.

\begin{figure*}[!t]
\centering
\subfloat[stickbot]{\includegraphics[width=1.1in]{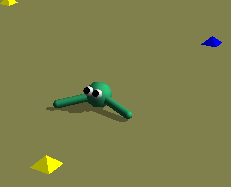}} 
\subfloat[twigbot]{\includegraphics[width=1.1in]{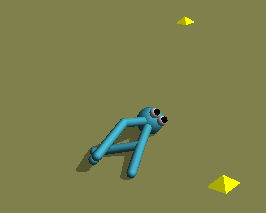}}
\subfloat[branchbot]{\includegraphics[width=1.1in]{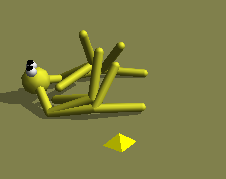}}
\subfloat[treebot]{\includegraphics[width=1.1in]{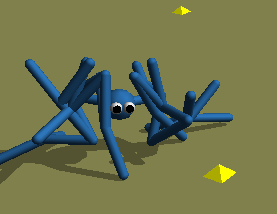}} 
\subfloat[spherebot]{\includegraphics[width=1.1in]{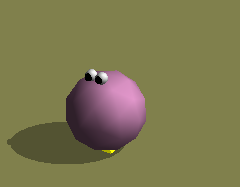}}\\
\subfloat[starfishbot]{\includegraphics[width=1.1in]{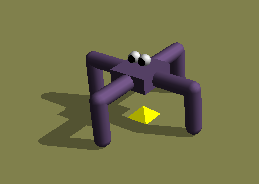}} 
\subfloat[crabbot]{\includegraphics[width=1.1in]{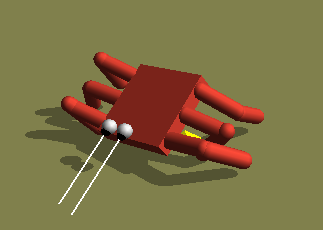}} 
\subfloat[quadruped]{\includegraphics[width=1.1in]{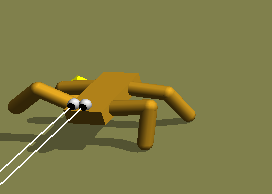}} 
\subfloat[tablebot]{\includegraphics[width=1.1in]{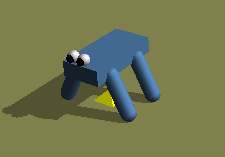}} 
\subfloat[snakebot]{\includegraphics[width=1.1in]{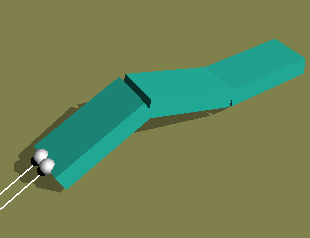}} 
\caption{The ten robot species employed in the experiment.}
\label{fig:species}
\end{figure*}

\section{Methodology}
Our method was compromised of four parts: 
the crowd proposed commands to simulated robots with differing morphologies (section~\ref{sec:morphology}); 
the robots acted and then received reinforcement under those commands (section~\ref{sec:commands}); 
the robots were evolved to collect as much positive reinforcement as possible (section~\ref{sec:evolution}); 
and critics were trained to predict the crowd's reinforcement of a given robot's action under a given command (section~\ref{sec:critic}). 

\subsection{Morphology} 
\label{sec:morphology}
We used ten morphologically disparate simulated robot species in our experiment, thus ensuring that their behaviors are all different (Fig.~\ref{fig:species}). 
The robots were simulated using Pyrosim\footnote{ccappelle.github.io/pyrosim/}, 
a Python Robot wrapper for 
Open Dynamics Engine physical simulator.

Of the 10 robot species, the stickbot, twigbot, branchbot and treebot are morphologically related in that they embody full binary trees with depths between 1 (stickbot) and 4 (treebot). 
A stickbot consists of two identical cylinders (segments), a twigbot six, a branchbot 14, and a treebot 30. 
Each child segment is connected to its parent segment by a 1-DOF hinge joint. 
The default angles between connected segment pairs can vary between robots within a species, making the members of the same species look and act differently.

The starfishbot, crabbot, tablebot, and the quadruped are morphologically similar. 
The starfishbot consists of eight cylinders (four legs), the crabbot twelve cylinders (six legs), the tablebot four cylinders (four legs), and the quadruped eight cylinders (four legs). 
The legs are connected to the main body with 1-DOF hinge joints. 
The hinge joints restrict the movement of quadrupeds and crabbots to sideways
{\color{black} motion} and the movement of tablebots to forward and backward {\color{black}motion}.   
The only two non-legged species are the spherebots and snakebots. 
A spherebot is made of a sphere with a pendulum inside connected to the center of the sphere by a 1-DOF hinge joint. 
A snakebot consists of three rectangular solids attached with two hinge joints such that its hinge joints restrict its movement to its sagittal plane.

\textbf{Sensors.} 
Each robot is equipped with one touch sensor in each body part, one proprioceptive sensor in each joint, a 3D position sensor in its head, and two distance sensors emanating from its eyes. 

\begin{figure}[!htb]
 \centering
 \includegraphics[width=0.25\textwidth]{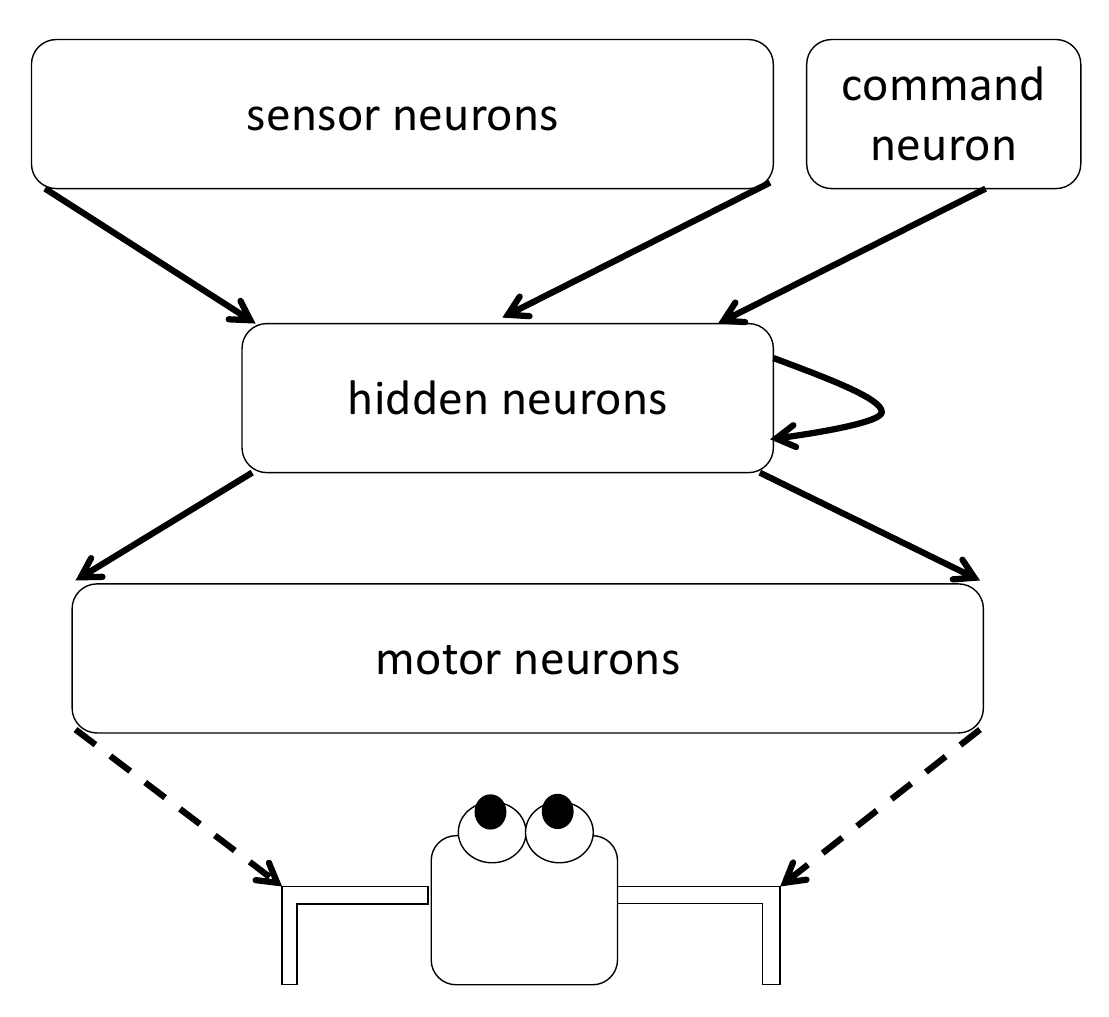}
 \caption{\label{fig:controller} The closed loop controller 
 {\color{black}
 } of a robot, where the input neurons are connected to both the robot's sensors and the {\color{black} command
 } neuron. The output neurons are connected to the robot's motor joints. The number of sensor (S) and motor neurons (M) for each robot species are: stickbot: (9S, 2M), twigbot: (17S, 6M), branchbot: (33S, 14M), treebot: (65S, 30M), spherebot: (6S, 1M), snakebot: (9S, 2M), tablebot: (13S, 4M), crabbot: (27S, 12M), quadruped: (21S, 8M), and starfishbot: (21S, 8M). 
}
\end{figure}

\textbf{Controllers.} 
The controller of each robot is a neural network with three layers as shown in Fig.~\ref{fig:controller}. 
The first layer contains one neuron for each sensor, one neuron for the command proposed by the crowd, and one bias neuron. 
The hidden layer contains five neurons with {\color{black}
} recurrent connections between each single and pair of neurons. 
Finally, the output layer contains one neuron for each motorized joint.  

\subsection{Commands and Reinforcement}
\label{sec:commands}

\begin{figure*}[!htb]
\centering
\includegraphics[width=1\textwidth]{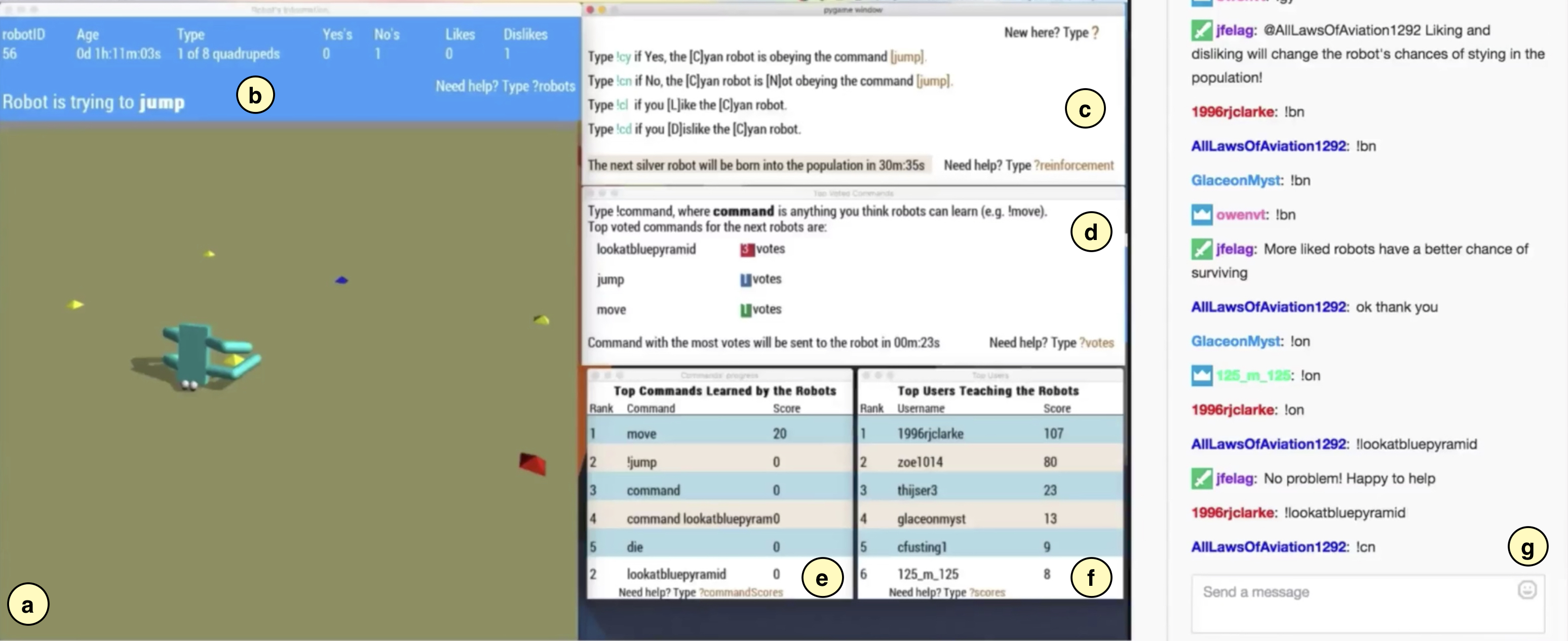}
\caption{A screen-shot of the TwitchPlaysRobotic channel. 
\textbf{a)} A robot is simulated for thirty seconds under the current command, \enquote{jump}. 
\textbf{b)} A panel listing information about the current robot including 
its ID, age, and type, 
the number of yes's and no's provided during the current evaluation, and 
the number of likes and dislikes the current robot has received overall.
\textbf{c)} A panel explaining how users can reinforce the current robot. 
\textbf{d)} A panel prompting users to propose or vote for the next command for the next three minute window. 
\textbf{e)} A top five commands by score list plus the current proposed command and score. 
\textbf{f)} A top five users by score list plus the last active user in the chat and their score. 
\textbf{g)} The live chat session where users enter reinforcement, commands, help inquiries, and other messages.}
\label{fig:front-end}
\end{figure*}

\textbf{Video streaming.} 
We employ Twitch.tv, a live video streaming platform, for users to watch and interact with the robots. 
Twitch.tv is particularly popular among people live streaming themselves playing video games as well as people who enjoy watching others play those games. 
Twitch.tv was previously used by \cite{anetsberger} for crowdsourcing. 
We used the Open Broadcaster Software (OBS) to broadcast the screen of the computer simulating the robots to the TwitchPlaysRobotics (TPR) channel on Twitch.tv, 
where users could watch them 
behave\footnote{{\color{black}twitch.tv/twitchplaysrobotics}}. 
Viewers can communicate with robots by entering commands and reinforcement into the live chat window (Fig.~\ref{fig:front-end}). 
The stream ran 24 hours a day for 31 days from July 18, 2017 to August 18, 2017. 

\textbf{Commands.} 
Every three minutes, the command that had been typed in most often is issued to each robot during the next three minute window. 
There were no pre-defined commands: users could vote for any command they could imagine by simply typing it in. 
If no commands were typed in during that three minute period, the next command would default to \enquote{move.} 
Each unique command was assigned a random real number in $(-1, +1)$, which is then supplied to one input neuron as shown in Figure~\ref{fig:controller}. 
This admittedly oversimplified method of language encoding was employed in the spirit of experimental simplicity. 

\textbf{Reinforcement.} 
Each robot is simulated for 30 seconds. During that time, users are prompted to provide positive or negative reinforcement of the action of the displayed robot under the current command, as well as whether they like or dislike the robot, by typing in `y', `n', `l', and `d', respectively. 
The users were instructed to preface each of these four characters with the first letter of the color of the robot in order to assign their reinforcement to the relevant robot.
Internet network latency can cause a vote to be given to the wrong robot if a user sees and reinforces robot $i$, but the broadcasting computer has already moved on to simulating robot $i+1$. 

\textbf{Incentivization.} 
In order to help users determine what commands are groundable by the robots, a table in the bottom right of the screen lists the top five commands (Figure \ref{fig:front-end}). 
A command's score is calculated as $C=(Y_{2}-N_{2})-(Y_{1}-N_{1})$, which measures whether ($C>0$) or not ($C<0$) this command is becoming grounded by the robots.  
$Y_{2}$ and $N_{2}$ represent the total positive and negative reinforcement received by robots under this command during the second half of the command's active period. 
A command's active period is defined as $t_T - t_0$, where $t_0$ denotes the time at which the command was first typed in, and $t_T$ denotes the last time it was typed in. 
Similarly, ${Y}_{1}$ and $N_{1}$ represent the total positive and negative reinforcement over the first half of a command's active period. 
The higher a command's score, the better all robots have evolved to obey the command. 
To incentivize users to participate, the top five users by score were also displayed. 
A user receives one positive point for each reinforcement or command they enter.

\subsection{Evolution}
\label{sec:evolution}
\begin{figure}[!tb]
 \centering
 \includegraphics[width=0.35\textwidth]{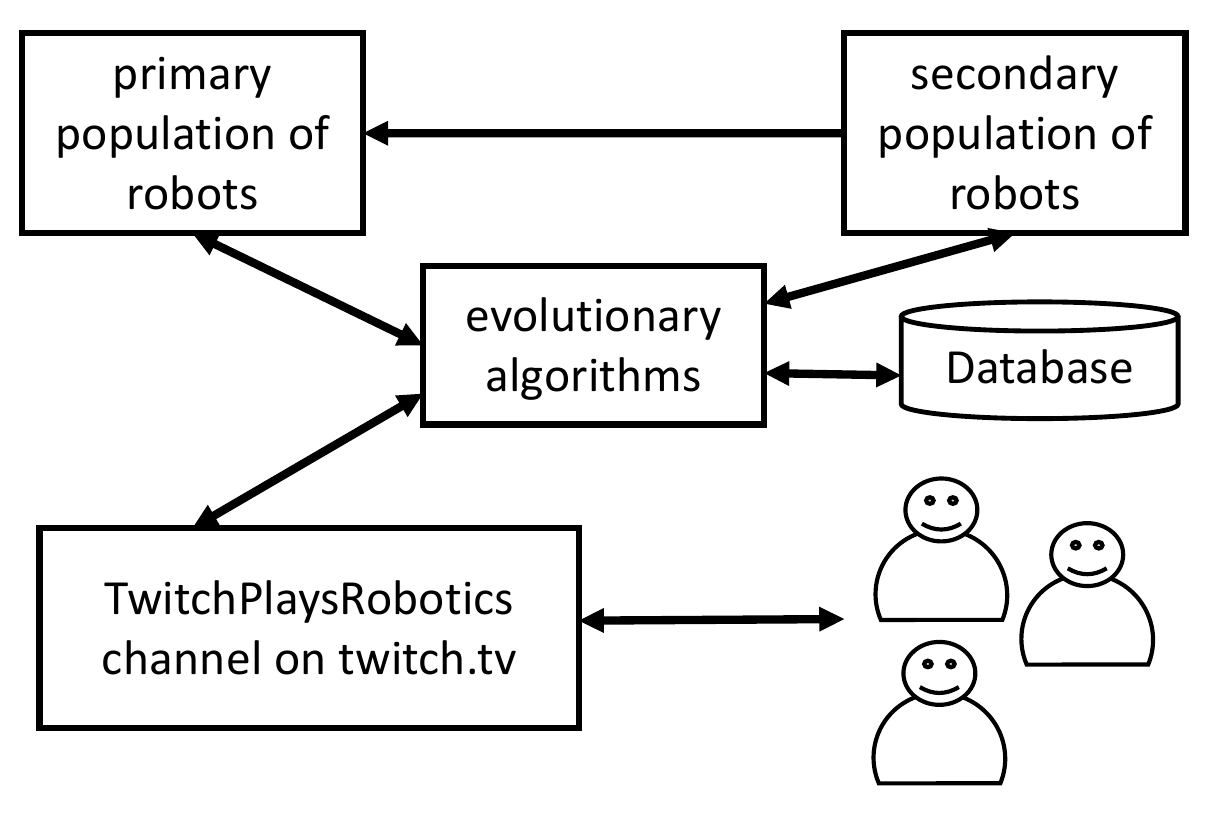}
 \caption{\label{fig:framework} A framework of different components including, Twitch.tv chat server, TwitchPlaysRobotics channel, primary and secondary populations of robots, MySQL database server, master program and the users interacting with each other. An arrow between any two components imply the two interact at some level. 
 }
\end{figure}

Figure~\ref{fig:framework} shows the main components of our system. 
The master program chooses which robot to show the crowd next and simulates it. 
Chat bots process incoming messages from the Twitch.tv chat server. 
The master program and chat bots store information about the users, their reinforcement and commands, and the simulated robots in a MySQL database. 
Two populations of robots were maintained throughout our experiment: 
a primary population, from which robots are selected and shown to the crowd; and a secondary population that injects new robots into the primary population. 

\textbf{Primary population.} 
The primary population begins with 50 robots, five from each of the 10 species (Fig.~\ref{fig:species}). 
Every 30 seconds, a randomly selected robot is simulated and drawn in 
red, green, blue, orange, cyan, or purple, and then back to red, with successive colors assigned to successive robots. 
This 30 second simulation of a robot with a given command will henceforth be referred to as an evaluation. 
During this period, the simulation was advanced $1800$ time steps, with a step size of $0.05$.
A robot may be selected and evaluated multiple times under the same command. 
While one robot is being evaluated, we choose two other robots at random from the primary population to compete. 
They do so based on three metrics: popularity (likes minus the dislikes), obedience (yes's minus no's), and number of evaluations. 
A robot dominates the other one if it achieves higher popularity and obedience with less or the same number of evaluations. 
If the two robots tie in popularity and obedience, then no change occurs. 
Otherwise, a mutated copy of the dominating robot replaces the dominated robot in the primary population. 
For all offspring, its neural network controller (or the mechanical structure in the case of the twigbots, stickbots, branchbots, and treebots) is mutated. 
 
\textbf{Secondary population.} 
The primary population was found to produce insufficiently interesting new behaviors on its own, so a secondary population was created to generate and inject new robot behaviors throughout the experiment: 
Users who observe continuously novel behaviors are more likely to continue participating in the experiment. 
The evolution of the secondary population was inspired by novelty search as reported in \cite{lehman}. 
The secondary population is partitioned into ten subpopulations, one for each species, with no interactions between the subpopulations. 
The secondary population is initialized with twenty random robots for each of the 10 species and evolved using a hill climber: 
each generation, each robot in the secondary population produces an offspring with a different body or neural controller, which replaces the parent if it is fitter.
Each robot is evaluated twice: 
once with the command neuron set to $+1$ and once 
{\color{black}with}
it set to $-1$. 
The fitness of a robot is set to the Euclidean distance between its trajectories under these two evaluations. 
This fitness function thus encourages the evolution of robots that react differently to different commands. 

At the outset of the experiment, we randomly remove five robots from each secondary subpopulation and assign those fifty robots as the initial primary population.
Every hour during the experiment, the secondary population cycles through 120 generations.
At the beginning of each hour, we randomly choose one of the 200 robots in the secondary population, inject it into the primary population, and replace it 
{\color{black}in the secondary population}
with a new robot with the same morphology and a random neural controller or body. 
The injected robot replaces the robot with the least evaluations in the primary population. 
If there is a tie between robots with the least evaluations, then one is chosen at random.
The injected robot is simulated and streamed to the crowd during the next evaluation window. 
It is colored silver, a color not in the color pool, to signal to the crowd that a new robot has been created.

\subsection{Predicting Crowd Response}
\label{sec:critic}
In this work, we define the safety of a robot as its ability to predict the crowd's reaction after it receives a command from them and simulates its own action in response to that command. 
This ability depends on how well a robot can ground the command that it hears in both its own sensorimotor experience and predicted social response. 
Therefore, after running the experiment for a month, multiple critics (as in the actor-critic paradigm) were trained, one for each robot, to learn relationships between commands, sensor data, and {\color{black}the}
crowd's reinforcement. 
We then examined whether different critics learned this relationship better or worse for different robots, indicating that some robots can be more safe than others, according to our definition. 

\textbf{Training data.} 
Training data for each critic was drawn from the interaction between the users and robots. 
First, all evaluations that failed to collect at least one positive or negative reinforcement were discarded. 
Among the reminder, the command \enquote{move} received more evaluations than the other commands over all the species. 
Most evaluations in this set received positive reinforcement. 
Thus, to balance the training dataset, we included the evaluations under the command \enquote{stop} by assuming that a robot which acts and receives no's under the \enquote{stop} command would have likely received yes's under the \enquote{move} command. 
Therefore, for each species, we created a dataset of all evaluations under those two commands. 
For each evaluation, each robot's sensors changed over $1800$ time steps. 
To train the critics, we employed the proprioceptive sensors for all the joints in the robot and the 3D coordinates of the head as recorded by the position sensor. 
To unify the number of features across species, we combined all of the proprioceptive sensors into one feature using $\tilde{j}_t = \frac{\sum_{t}\sum_{i}(j_{it} - j_{i(t-1)})}{n_t}$, where $j_t$ is a robot's $i$th proprioceptive sensor value at time $t$ and $n$ is the total number of proprioceptive sensors. 
This feature thus represents the average change in the joints across the robot, throughout its evaluation period.

Next, to reduce the number of features further, the number of time steps were reduced from $1800$ to $100$ by retaining every 18th sensor value, for all sensors, and discarding the rest. 
We also normalized the reinforcement of each evaluation to the range of $[-1,+1]$ using $o_i = \frac{|e_{iy}| - |e_{in}|}{|e_{iy}| + |e_{in}|}$ where $|e_{iy}|$ and $|e_{in}|$ represent the number of positive and negative reinforcement that the evaluation $i$ received, respectively. 
We assumed that a robot that receives a negative normalized reinforcement ($o_i<0$) for \enquote{stop} is likely to receive a positive normalized reinforcement ($o_i>0$) for \enquote{move} and vice versa, so we invert the sign of the normalized reinforcement for evaluations under the command \enquote{stop}. 
After this feature processing stage, each critic receives a $100 \times 4$ matrix for each evaluation and outputs a scalar value. 
A dataset of $200$ of these input/output pairs that contains 100 positive example{\color{black}s} ($o=+1$) and 100 negative example{\color{black}s} ($o=-1$) {\color{black}was} 
prepared for {\color{black} eight
} of the 10 species
{\color{black}(all 10 species received differing amounts of such evaluations, but all received at least 200)}. 
The twigbot and stickbot did not receive enough evaluations with $o>0$, so we were not able to train critics for them.  

\begin{figure}[!t]
 \centering
 \includegraphics[width=0.2\textwidth]{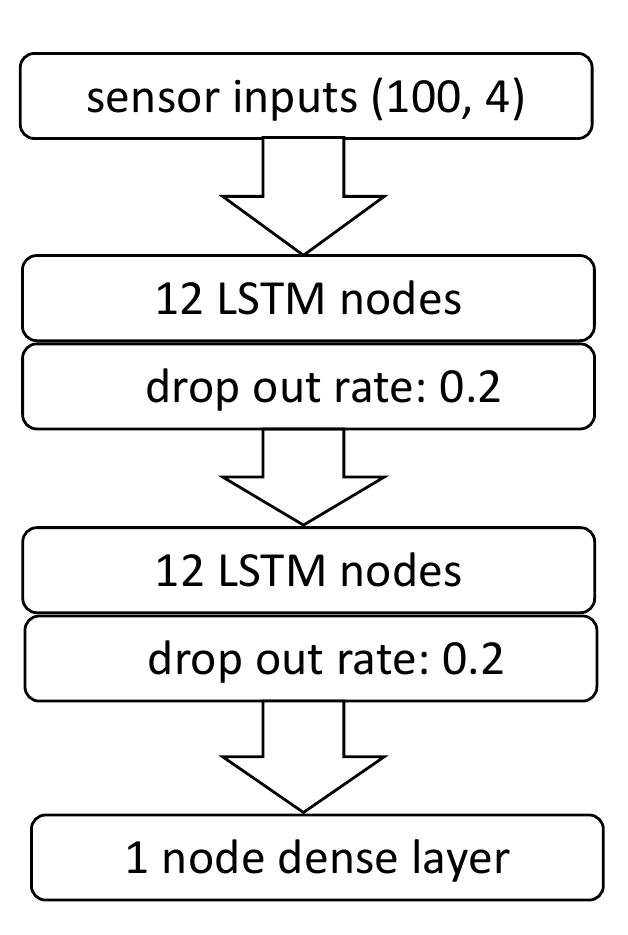}
 \caption{\label{fig:model}The critic, a Recurrent Neural Network, was trained to predict the crowd's actual response.}
\end{figure}

We implemented the critics in Keras (\cite{chollet2015keras}), a high-level neural network API in Python with TensorFlow (\cite{tensorflow}) as the backend. 
Each critic, as shown in Fig.~\ref{fig:model}, features two Recurrent Neural Network (RNN) stacks with twelve Long Short Time Memory (LSTM) cells in each RNN stack. 
Each stack was followed by a dropout layer with a dropout rate of $0.2$. 
The last layer contains one neuron that receives connections from each neuron in the previous layer. 
Each was trained and tested with 30 fold cross-validation and 100 epochs. 

\section{Results}
\subsection{Crowd Deployment}
During the month of deployment, 550 users issued 516 unique commands and 17,229 total reinforcement signals: 12,247 positive or negative reinforcements and 4,982 likes or dislikes. 
Over 2000 robots were either born into the primary population or injected into it from the secondary population. 

Besides the quantitative results of the deployment, it is interesting to explore the social interaction among users. 
Some of the social interactions observed include, but are not limited to, peer correction, social agreement (or disagreement) about reinforcement, and social copying. 
An example, listed in Table \ref{tab:disagreement}, shows two users discussing whether the displayed robot obeyed the current command \enquote{run}.

\subsection{Model Prediction}
\begin{figure}[!t]
\centering
\includegraphics[width=0.5\textwidth]{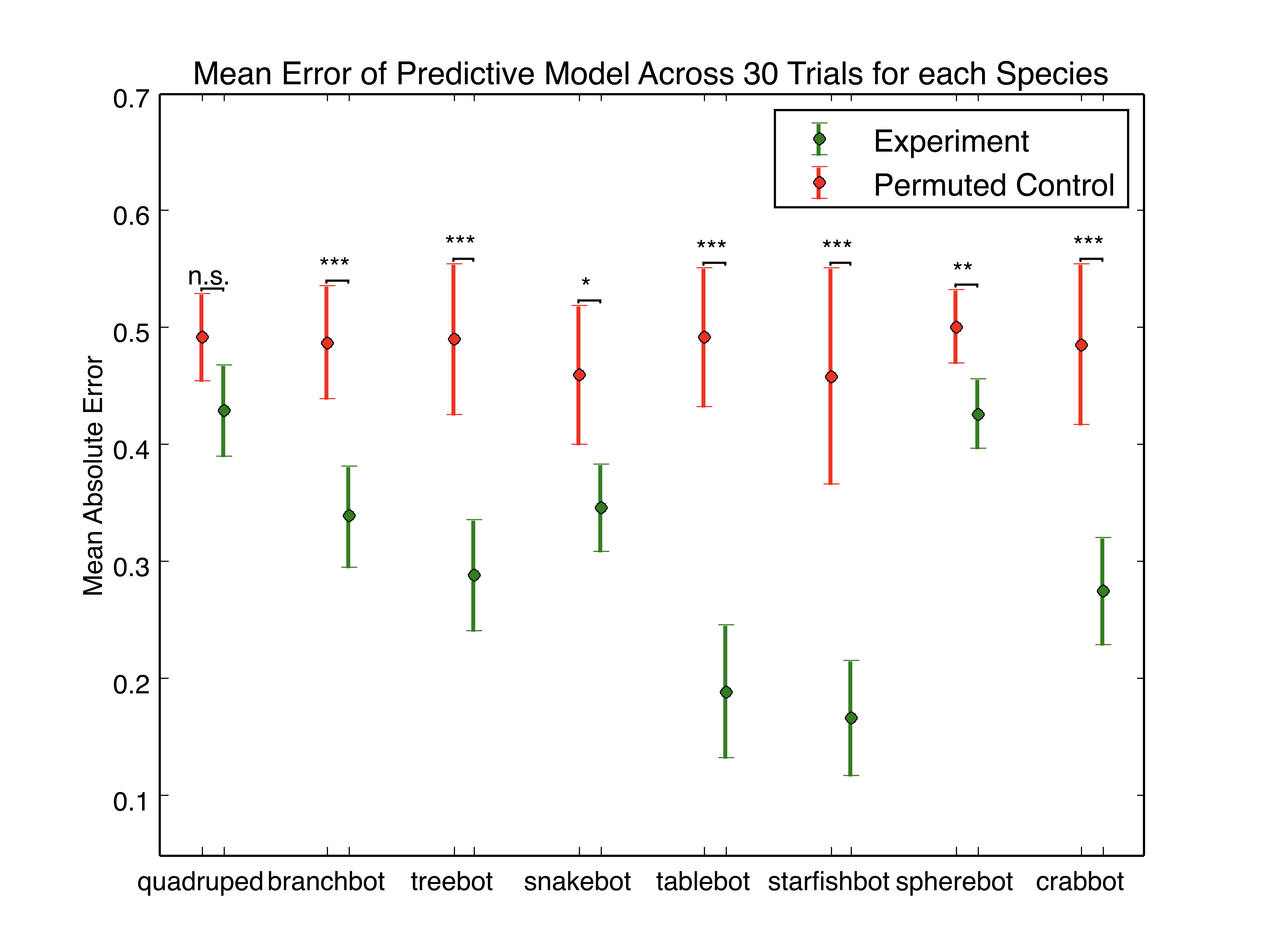}
\caption{\label{fig:critic_result} Prediction results for eight species under command \enquote{move} and \enquote{stop}. The $p$-value between the experimental treatment of the starfishbot and others are: quadruped=***, branchbot=***, treebot=**, snakebot=***, tablebot=n.s., spherebot=***, crabbot=*.  The $p$-value between the experimental treatment of the tablebot and others are: quadruped=***, branchbot=***, treebot=n.s., snakebot=***, starfishbot=n.s., spherebot=***, crabbot=n.s. (***=$p<0.001$, **=$p<0.01$, *=$p<0.05$, and n.s.=otherwise). Error bars report a $99\%$ confidence interval.}
\end{figure}

The mean error for the 30 critics, when exposed to their respective testing sets, was calculated as $M_e = \frac{\sum_{i=1}^{N}{(o^{\prime}_i - o_i)}}{n}$, where $n$ denotes the number of samples in a test set and $o^{\prime}_i$ denotes the predicted normalized reinforcement (labeled \enquote{experiment} in Fig.~\ref{fig:critic_result}). 

As {\color{black} a} control, the critics were exposed to the same testing sets but with the reinforcement signals randomly permuted (Fig.~\ref{fig:critic_result}). 
This permuted control was designed to ensure that users did not provide random reinforcement (in which case both treatments would yield equally high error) or unanimous positive or negative reinforcement (in which case both treatments would yield equally 
{\color{black} low
} 
error). 
The \textit{p}-value between the experimental and permuted treatment was calculated with Student's \textit{t}-test and multiplied by the Bonferroni correction of ${\binom{16}{2}}$ for multiple comparisons. 

Figure~\ref{fig:critic_result} shows that the prediction error for the quadruped was not significantly lower than its corresponding control experiment.
The other robot species however show that their critics can predict the crowd's response based on self-sensed features of their behavior under this pair of commands. 
A similar observation was also reported in the work conducted by \cite{anetsberger} for different robots and a different command, confirming that unpaid participants can and do provide sufficient quantity and quality of reinforcement to yield safe, yet simple robots. 
It is important to note that the prediction error is still relatively high for the regular experiments. 
However, 
{\color{black}lower
}
prediction errors may be achieved with larger training datasets, a different critic architecture, or different hyperparameters of the critics.

In addition to grounding \enquote{move} and \enquote{stop}, it can be seen from the mean error of the critics that the starfishbot and tablebot achieved a significantly lower average error compared to some of the other species. 
More specifically, the critic of tablebot was significantly more accurate than 
{\color{black}the other critics
}
except 
{\color{black}for the critics of}
the treebot, crabbot, and starfishbot.
The critic of the starfishbot was more accurate than 
{\color{black}the critics for the other species
}
except the tablebot.
The \textit{p}-value between the experimental treatments of 
{\color{black}this pair of species against the other six species
}
was calculated with Student's \textit{t}-test and multiplied by the Bonferroni correction of ${\binom{16}{2}}$ for multiple comparisons. 
Thus, these two species have been able to ground the two symbols more accurately.
This suggests that {\color{black}
} morphology has an impact on the way the robots grounded these symbols. 
One possible reason for this result could be that the crowd was more certain about the obedience (or disobedience) of these two species, under these two commands, and thus provided unanimous reinforcement more often for them than they did for the other species. 
It is possible that the crowd could be more or less certain about the obedience of a robot because of its range of motions or the way moves in the environment, or visual obstruction of parts of the body. 
However, none of these explanations for the observed difference in safety have yet been verified.

\section{Conclusion and Future Work}
Here, we show that morphology may affect the crowd's ability to render a robot safe, where safety is defined as the robot's ability to predict human reactions to an action it performs in an attempt to obey a proposed command. 
Although the morphological properties that make a particular robot safer 
than others are not yet understood, 
{\color{black}this work suggests that}
it is 
{\color{black}important
}
to design future robots with morphology in mind.

{\color{black}The robotics community could benefit from the present work in the following manner.
During the design stage of a given robot, mechanically-different variants of it could be simulated in virtual environments, streamed to a web service such as Twitch.tv, and observers could be told what task the robot should perform. Controllers could then be optimized for these variants using crowd reinforcement. If controllers can be trained on a variant to consistently elicit positive reinforcement, then that prototype is capable; if critics can also be trained for the same variant to successfully predict the crowd’s responses, it is also safe. After a physical version of this capable and safe robot is manufactured, fitted with the trained controller and critic, and deployed, the controller and critic of the physical robot can continually be adjusted to any unforeseen changes through continuous simulations in parallel with reality.}
Connecting a physical robot to a simulator is beyond the scope of the current experiment, 
but \cite{bongard2006} and \cite{cully2015} studied how a physical robot can generate a model of the environment and its self, suggesting this might be a possible direction of future study. 

Although we used evolution to train robots to ground as many commands as possible, we did not observe the robots evolve to obey increasing numbers of the proposed commands. 
This problem is mainly due to catastrophic forgetting: a population of robots might have evolved to obey a particular command, but the next command proposed by the crowd would have likely killed them. 
Thus, we wish to employ more sophisticated objective functions and search algorithms to avoid or minimize catastrophic forgetting
{\color{black} in future}. 
Another factor that might have contributed to lack of evolution is that we tried to train controllers to produce different behaviors using only one neuron for a command. 
In the next deployment, we plan to use word2vec to encode the crowd's proposed commands as input to the robots' controllers.

\begin{table}
\centering
\begin{tabular}{|lll|}
\hline
 time & username & message \\\hline
2017-08-08 10:45:43 & senorpieg & !bn\\
2017-08-08 10:45:45 & jackcq395 & !by \\
2017-08-08 10:45:46 & senorpieg & not running \\
2017-08-08 10:45:48 & senorpieg & walking \\
2017-08-08 10:45:51 & senorpieg & difference \\
2017-08-08 10:45:55 & senorpieg & closer \\
2017-08-08 10:45:56 & jackcq395 & thats how i run \\
2017-08-08 10:45:59 & senorpieg & but difference \\\hline
\end{tabular}
\caption{Example of disagreement among users about a blue robot evaluated under the command \enquote{run}.\label{tab:disagreement}}
\end{table}

\section*{Acknowledgments}
This work was supported by the National Science Foundation under grant No. EAGER-1649175.

\bibliographystyle{named.bst}
\bibliography{ijcai18}

\end{document}